\documentclass[a4paper, 10pt, conference]{ieeeconf}      
\usepackage{FG2024}
\usepackage[nolist]{acronym}
\usepackage{url}
\usepackage{hyperref}
\acrodef{SL}[SL]{sign language}
\acrodef{SLR}[SLR]{sign language recognition}
\acrodef{CNN}[CNN]{convolutional neural network}
\acrodef{ISLR}[ISLR]{Isolated sign language recognition}
\acrodef{BSign22k}[BSign22k]{BosphorusSign22k}
\acrodef{AUTSL}[AUTSL]{Ankara University Turkish Sign Language Dataset}
\acrodef{MCC}[MCC]{Minimum Class Confusion}
\acrodef{DANN}[DANN]{Domain Adversarial Neural Network}
\acrodef{JAN}[JAN]{Joint Adaptation Network}
\acrodef{MDD}[MDD]{Margin Disparity Discrepancy}
\acrodef{DSBN}[DSBN]{Domain Specific Batch Normalization}
\acrodef{DGS}[DGS]{German Sign Language}
\acrodef{ASL}[ASL]{American Sign Language}
\acrodef{BSL}[BSL]{British Sign Language}
\acrodef{TID}[TID]{Turkish Sign Language}
\acrodef{SL-GCN}[SL-GCN]{sign language graph convolution network}

\FGfinalcopy 

\IEEEoverridecommandlockouts                              
\usepackage{graphicx} 
\usepackage{booktabs}

\def\FGPaperID{****} 

\title{\LARGE \bf
Transfer Learning for Cross-dataset Isolated Sign Language Recognition in Under-Resourced Datasets
}

\author{\parbox{16cm}{\centering
    {\large Ahmet Alp Kindiroglu$^{1,2,}$\textsuperscript{*}, Ozgur Kara$^{3,}$\textsuperscript{*}, Ogulcan Ozdemir$^1$ and Lale Akarun$^1$}\\
    {\normalsize
    $^1$ Department of Computer Engineering, Bogazici University, Istanbul , Turkey\\
    $^2$ Huawei Turkey R\&D Center, Istanbul , Turkey\\
    $^3$ Georgia Institute of Technology, USA}}
    \thanks{The numerical calculations reported in this paper were partially performed at TUBITAK ULAKBIM, High Performance and Grid Computing Center (TRUBA resources).}
}

\usepackage{fancyhdr}
\begin{document}

\ifFGfinal
\thispagestyle{empty}
\pagestyle{empty}
\else
\author{Anonymous FG2024 submission\\ Paper ID \FGPaperID \\}
\pagestyle{plain}
\fi
\maketitle

\begingroup\renewcommand\thefootnote{*}
\footnotetext{Equal contribution}
\endgroup

 \thispagestyle{fancy} 
\begin{abstract}

Sign language recognition (SLR) has recently achieved a breakthrough in performance thanks to deep neural networks trained on large annotated sign datasets. Of the many different sign languages, these annotated datasets are only available for a select few. Since acquiring gloss-level labels on sign language videos is difficult, learning by transferring knowledge from existing annotated sources is useful for recognition in under-resourced sign languages. This study provides a publicly available cross-dataset transfer learning benchmark from two existing public Turkish SLR datasets. We use a temporal graph convolution-based sign language recognition approach to evaluate five supervised transfer learning approaches and experiment with closed-set and partial-set cross-dataset transfer learning. Experiments demonstrate that improvement over finetuning based transfer learning is possible with specialised supervised transfer learning methods. 

\end{abstract}

\section{Introduction}
\label{sec:intro}

\Acp{SL} are visual languages that use movements and expressions of hands, arms, and faces to convey meaning. Having developed naturally out of local deaf communities, more than 103  \acp{SL} are known to exist and show considerable variability from one another \cite{Woll2001}.

Isolated \ac{SLR} from sign language videos is an active research topic. In recent years, deep learning methods have shown great success in challenging video classification problems such as activity recognition owing to the emergence of huge annotated datasets. Similar high recognition accuracies were observed in isolated \ac{SLR} for some large annotated public datasets \cite{jiang2021skeleton, bohavcek2022sign,selvaraj2021openhands}. For sign languages, annotating videos is a time-consuming task. Glosses, which are transcribed approximations of signs in spoken languages, are annotated by expert signers in the order they appear in the video. This difficult task consumes 10-30 minutes for a minute-long sign video. While these annotation efforts exist for some languages such as \ac{DGS} \cite{jahn2018publishing}, \ac{BSL} \cite{Albanie2021bobsl} or \ac{ASL}  \cite{neidle2012challenges}, most of the remaining languages are still considered  under-resourced languages. In order to train successfully working isolated \ac{SLR} models with such languages, the most commonly used approach is to utilize transfer learning from similar datasets or tasks to improve recognition accuracy.

Transfer learning approaches aim to use data from data-rich sources to improve task performance on a data-poor target task. For isolated \ac{SLR}, our focus is on \ac{TID}, where two medium-scale public datasets exist. These are the \ac{BSign22k} \cite{ozdemir2020bosphorussign22k} dataset with 22k videos for 745 distinct signs and the \ac{AUTSL} \cite{sincan2021using} dataset with 32k videos for 216 signs. Compared to common video classification datasets such as Kinetics \cite{kay2017kinetics}, or MomentsInTime \cite{monfort2019moments} with thousands of samples per class, these datasets, along with most other isolated \ac{SLR} datasets, can be considered as under-resourced datasets. 

It is reported that transferring knowledge from larger datasets often improves accuracy, while a little benefit is observed when a transfer is attempted between two similarly sized datasets \cite{selvaraj2021openhands,  jiang2021skeleton}. Although transfer learning datasets exist for video classification \cite{chen2019temporal}, there is no SL transfer dataset to test different transfer strategies.

In this paper, we establish a sign language recognition baseline for cross-dataset sign language recognition using two public isolated \ac{TID} datasets. These datasets share 57 common signs. However, not all signs are identical, as their performance varies from dataset to dataset due to differences in interpretation and style. Cross-dataset sign language recognition can approximate and measure such differences in sign and signer characteristics. Among these two datasets, \ac{BSign22k} contains fewer samples per class, so transfer learning from \ac{AUTSL} to \ac{BSign22k} is chosen as the use case that makes more sense. Using this setting, we experiment with two different transfer learning settings. We construct a transfer learning pipeline using the \ac{SL-GCN} algorithm as our baseline feature extractor.

Existing transfer learning methods make use of combined training and finetuning based methods \cite{jiang2021skeleton, selvaraj2021openhands}. In contrast, to transfer signs between two supervised source domains, we use five different transfer learning techniques, namely, finetuning, \ac{DANN}, \ac{MCC}, \ac{JAN} and \ac{DSBN} under two settings. In setting 1, common signs between source and target datasets are unknown. In setting 2, a closed set transfer learning problem is observed where source and target class labels are identical and known. Setting 3 is a partial set transfer learning problem where the target set contains a subset of class labels belonging to the source set. In all settings, we try to find transfer learning approaches that yield the largest improvement compared to baseline approaches.  
The contributions of this study can be summarized as follows:

\begin{itemize}
    \item We propose a shared vocabulary sign language subset from two publicly available \Acf{TID} datasets on which supervised transfer learning approaches can be tested. \cite{sincan2020autsl,ozdemir2020bosphorussign22k}.
    
    \item Based on the \ac{SL-GCN} based sign language recognition model proposed by \cite{jiang2021skeleton}, we propose a baseline transfer learning experiment protocol. We use Minimum Class Confusion (MCC), Domain Adversarial Neural Network (DANN), Joint Adaptation Network (JAN) and Domain Specific Batch Normalization (DSBN as well as finetuning approaches. Results are presented for closed set and partial-set transfer learning cases.
    
    \item Signer Independent experiments using  MCC, DANN, JAN and DSBN show that transferring cross-dataset class knowledge with the proposed approach outperforms baseline finetuning and combined training approaches. Our work will be a baseline study for researchers who are working on under-resourced sign language recognition and video classification tasks.
\end{itemize}

This paper is organized as follows: In Section~\ref{sec:related_work}, we review the sign language literature. The implemented transfer learning methods, and our \ac{SL-GCN}-based sign language recognition framework is presented in Section~\ref{sec:technical}. Section~\ref{sec:experiments} describes the proposed dataset as well as the experiments and analysis. Finally, conclude the paper in Section \ref{sec:conclusion}.

\section{Related Work}
\label{sec:related_work}

Isolated \ac{SLR} involves the task of recognizing performed signs from a controlled vocabulary of signs in short video snippets. In the recent decade, several large isolated \ac{SLR} datasets such as Devisign \cite{chai2014devisign}, BosphorusSign22k \cite{ozdemir2020bosphorussign22k}, WL-ASL \cite{li2020word}, AUTSL \cite{sincan2020autsl}, CSL \cite{huang2018attention}, MS-ASL \cite{joze2018ms} have been collected. Sign language recognition may be performed using different modalities from these datasets, such as RGB frames, depth images, pose information, and motion flow. Among the RGB based state-of-the-art methods, 3D convolution-based methods such as I3D \cite{joze2018ms,li2020word}, 3D Resnets \cite{gokcce2020score} or temporal models that model frame-level features such as HMMs \cite{koller2017re,AZAR2020101053} or LSTMs \cite{sincan2019isolated,Saleh2020} are popular. In this study, we focus on pose-based Sign Language Recognition methods. There exist a variety of pose-based methods that can model sign languages as good as RGB-based recognition approaches. These methods often use pose information extracted through popular open-source pose extraction libraries that can extract hand joints such as OpenPose \cite{cao2019openpose}, MMpose\cite{mmpose2020} or MediaPipe Holistic\cite{grishchenko2020mediapipe}. In isolated \ac{SLR} there are pose based methods that print coordinate locations on images to recognize them with 2D Convolutions. Among these, SSTCN \cite{jiang2021skeleton} uses temporal convolution networks and Temporal Accumulative Features \cite{kindiroglu2019temporal} constructs motion energy features from hand and body joints.
In contrast, methods that model coordinates directly have recently become popular. \cite{joze2018ms} uses hierarchical co-occurrence networks(HCN) to model signs, \cite{bohavcek2022sign} uses transformers, while several other works use Graph Convolutional Networks(GCN). One GCN based method that is currently state-of-the-art in \ac{SLR} is \ac{SL-GCN} \cite{jiang2021skeleton, selvaraj2021openhands}.  It is based on the Spatio-temporal graph convolutional networks (ST-GCN) method from the action recognition domain, that combines temporal convolutional neural networks with spatial graph based convolutions of joints \cite{yan2018spatial, de2019spatial}. A method called drop-graph \cite{cheng2020decoupling} enhances the ST-GCN method by using an attention-based node drop mechanism called drop-graph for regularization. In isolated \ac{SLR}, \ac{SL-GCN} is used \cite{jiang2021skeleton, selvaraj2021openhands} by further adding spatial and temporal attention mechanisms to each layer.  Variations of the method such as  \cite{al2022spatial} have shown promising \ac{SLR} performance on other datasets using 3DGCN layers with a spatial attention mechanism. Other studies on improving graph convolution networks include Modulated GCN where weights and node affinities are modulated to optimize edges between nodes beyond skeletal connections \cite{zou2021modulated}. GCN based methods were also extended to continuous sign language recognition \cite{parelli2022spatio}, and have shown promising results on datasets such as  RWTH-PHOENIX Weather.


Research on isolated \ac{SLR} with the current state-of-the-art video classification methods yields high recognition accuracies when there is a sufficiently large dataset \cite{selvaraj2021openhands}.Transfer learning approaches may make it possible to carry this success to under-resourced sign languages where large annotated datasets are not available. Transfer Learning is a machine learning approach that utilizes labeled data from relevant source domains to train a model in the target domain \cite{wang2018deep}. In this work, we focus on supervised cross-dataset transfer learning where the performed tasks are similar, but datasets show a difference in data distributions, label-space differences, or in the amount of available training data samples per class. Deep domain adaptation methods aim to create representations that are robust to transfer by embedding domain adaptation in the pipeline of deep learning \cite{ganin2016domain, long2017deep, jin2020minimum}. In most of these methods, an encoder and a classifier layer are the primary deep learning model components whose weights are shared for both domains. Various distribution distance minimization approaches such as loss functions \cite{tzeng2014deep}, layer mechanisms \cite{chang2019domain}, and optimization functions are developed to promote domain confusion in the feature space. One line of research uses maximum mean discrepancy (MMD) \cite{tzeng2014deep} and correlation alignment \cite{sun2016deep} to reduce distribution shift between models learned from different sources. One problem with these approaches is that distribution alignment of the data does not automatically lead to the semantic alignment of different domains. Another line of research uses domain-specific weights to maximize target domain performance. Bousmalis et al. \cite{bousmalis2016domain} jointly learn the domain-shared encoder and domain-specific private encoders with domain separation networks. Chang et al. \cite{chang2019domain} create domain specific batch normalization layers for the feature extractor model while sharing all other model components. Recent studies such as \ac{MCC} \cite{jin2020minimum} and Batch nuclear-norm maximization \cite{Cui_2020_CVPR} have shown that loss functions based on regularization of unlabeled data can improve transfer learning performance without modifying the base feature extractor of the model.

Compared to image-based domain adaptation (DA),  Video-based DA attracted fewer studies. In video classification, models have to take into account the temporal variations on top of variations in the image space. Only a few works focus on small-scale video DA with only a few overlapping categories \cite{chen2019temporal, xu2021aligning}. Several methods aim to use image-based domain adaptation methods with 3D classification networks, such as adding a gradient reversal layer for domain invariance \cite{bellitto2021hierarchical}. The TA3N study\cite{chen2019temporal}  proposes a method using domain-specific attention while learning to align frames across domains. They demonstrate the improvement in performance by introducing the UCF-HMDB\textsubscript{full} and Kinetics-Gameplay datasets, providing a benchmark for cross-dataset transfer learning in action recognition.

In isolated \ac{SLR}, the use of transfer learning can be seen in cross-task transfer (from image recognition, action recognition \cite{Sarhan2020}, pose estimation \cite{Albanie2020} or continuous SLR \cite{Li2020}) or cross-lingual \cite{selvaraj2021openhands}  settings. However, as noted in several studies such as \cite{selvaraj2021openhands}, finetuning-based cross-dataset transfer between similarly sized isolated \ac{SLR} datasets of different sign languages shows little to no benefit. In order to understand what information can be transferred in isolated \ac{SLR} successfully with which method, a dataset that would allow us to experiment with closed and partial set transfer learning settings was necessary. Using such a dataset, it would become easier to independently test cross-dataset sign language transfer using several state-of-the-art transfer learning approaches. Such datasets exist for similar domains such as image classification \cite{Ringwald_2021_WACV}, video classification \cite{chen2019temporal}, but not sign language recognition. As such a dataset existed for video classification  with the HMDB \cite{kuehne2011hmdb} and UTF-101\cite{soomro2012ucf101} datasets, we follow similar procedures to construct a baseline subset of the \ac{BSign22k} and \ac{AUTSL} datasets.

\section{Technical Approach}
\label{sec:technical}

\subsection{SL-GCN based Isolated Sign Language Recognition}

In this section, we describe the individual components of our isolated \ac{SLR} pipeline. We perform coordinate-based Sign Language Recognition using coordinates extracted with the OpenPose library \cite{cao2019openpose} from each dataset. 

 In our isolated sign language recognition datasets, we have a set of videos composed of varying lengths of videos. Each video contains an RGB component containing a single user and a set of coordinates belonging to different joints of that user. Each coordinate consists of a triplet of values: the X and Y coordinates within the image, as well as the confidence level of joint detection for each timestep. We obtain $J=30$ joints belonging to fingertips, finger bases, wrists, arms, neck, mouth, nose, and eyes for each frame of a sign language video.

Having obtained the joints for each sign, we apply several normalizations and augmentations to make our models more robust to small changes in user performance. A typical property of the isolated \ac{SLR} datasets we use is that they both share the same rest pose where hands rest to the side of the users' legs. Frames that contain stationary hands in this rest pose are trimmed from the beginning and end of each video segment. In addition, further sampling is done from the sampled frames to bring the number of total frames sampled from each video to $J$ frames. This was done using a random uniform sampling from a set of fixed intervals. This approach created duplicates of frames for shorter clips and provided temporal variation when sampling from longer clips.

For each coordinate, we apply a spatial normalization where the origin of the 2D coordinate system is moved to the temporal mean of the neck joint for that sign. In addition, random horizontal mirroring and random spatial coordinate translation augmentations were found to improve recognition performance and were thus added to our pipeline.

The classifier base of our model is the \ac{SL-GCN} model proposed by Jiang et al.~\cite{jiang2021skeleton}. The model takes as input a sequence of fixed length coordinates and outputs class prediction probabilities for sign language gestures. 

We model the problem so that nodes of the graph correspond to joint landmark locations. The spatio-temporal graph adjacency matrix $A$ is constructed in the spatial domain according to anatomical spatial ordering, where neighboring joints are assigned a value of 1 and all other joints are assigned a value of 0. In the temporal domain, all the joints are only connected to themselves.   

The architecture of the model consists of ten \ac{SL-GCN} blocks for node and edge processing. Each \ac{SL-GCN} block consists of a spatial convolution layer, multiplication with the adjacency matrix, and temporal convolution layer. The proposed architecture can be seen in Figure 1. 

\begin{figure}[!htpb]
\label{fig:base}
\begin{center}
\includegraphics[width=0.9\linewidth]{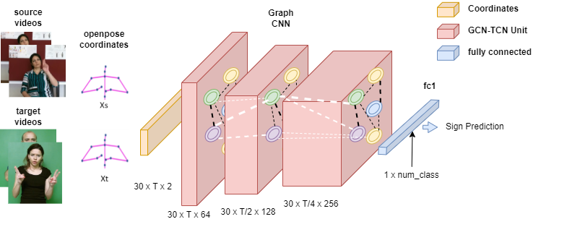}
\caption{Representative Image of the proposed SL-GCN architecuture. The proposed baseline method takes as input coordinates and outputs classification results for each video.}
\end{center}
\end{figure} 

The system uses the video modality to obtain coordinates that provide joint information for both hands. As shown in the figure, the system obtains coordinates from the OpenPose library, which is then used for training. During training, target domain videos are used, and in transfer learning settings, source domain videos are also added to the training process by adding an equal number of videos from each source domain to each batch. The baseline model is then trained using cross entrophy loss to perform classification.

The model includes decoupling with drop graph module proposed in \cite{cheng2020decoupling} and spatio temporal channel (STC) attention modules proposed in \cite{shi2020skeleton}. The decoupling adds increased recognition power by dropping random joints along with their neighbors closer than an adjacency of $K$. Similarly, the STC attention module increases recognition power by focusing on important joints, frames, and coordinates for certain signs. 

\subsection{Supervised Transfer Methods}
For each different transfer learning setup, this baseline feature extractor is used with different loss functions and classifiers in our experiments. We use several different transfer learning methods that aim to perform transfer learning by focusing on different characteristics of the data. 

The \textbf{{\acf{DANN} }} approach proposed in \cite{ganin2016domain} promotes the use of a gradient reversal layer to learn domain-independent features. In transfer learning setting, the model takes half the samples in each mini-batch from the source and target domains. The model has two classifiers, one for class prediction and the other for dataset prediction. The model learns to make accurate label predictions from samples of both datasets while adversarially trying to suppress any features that can allow the model to distinguish between samples of data from different datasets, such as user physical or performance-related characteristics. 

\begin{figure}[!htpb]
\label{fig:base}
\begin{center}
\includegraphics[width=0.9\linewidth]{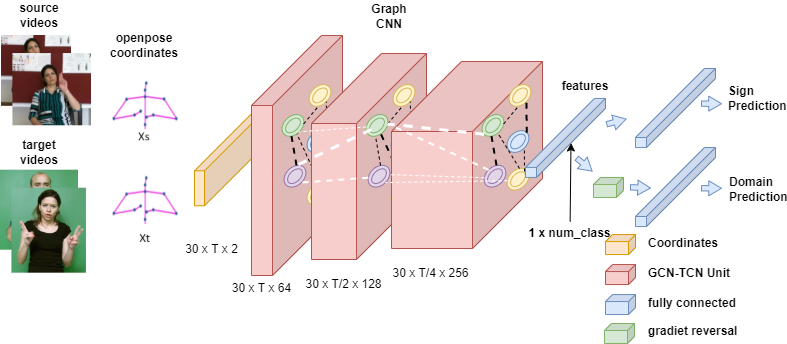}
\caption{Using gradient reversal on domain prediction loss, sign language recognition accuracy across source and target domains are improved. }
\end{center}
\end{figure}

The \textbf{{\acf{DSBN}}} approach shares all model parameters except batch normalization layers within the feature encoder. The batch normalization layer in a neural network regularizes feature representations from different domains without taking into account class or domain information. However, when domain discrepancy is significant, the effect of batch normalization is diminished. In this approach, individual batch normalization layers keep track of unique normalization parameters and batch statistics for each domain as seen in Figure 3. 

\begin{figure}[!htpb]
\label{fig:base}
\begin{center}
\includegraphics[width=0.9\linewidth]{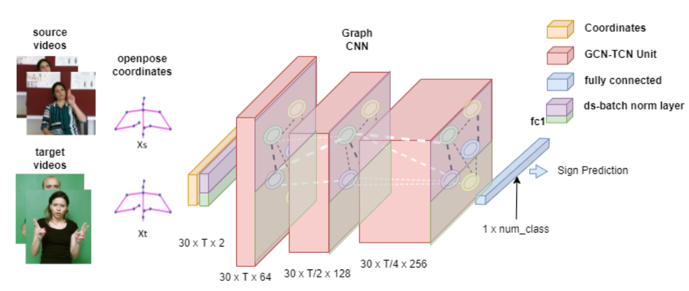}%
\caption{Domain specific batch normalization layers are learned to perform batch specific normalization. }
\end{center}
\end{figure} 

The \textbf{{\acf{JAN}}} approach proposed in \cite{long2017deep} maps features from different domains into a new data space where inter-class features have a more significant similarity. The method proposed, named joint maximum mean discrepancy (JMMD), minimizes the joint probability distribution distance of the source and target class-specific layers. The approach adds task-specific layers on top of the base \ac{SL-GCN} network to learn mapping to a common domain as seen in Figure 4.

\begin{figure}[!htpb]
\label{fig:jan}
\begin{center}
\includegraphics[width=0.9\linewidth]{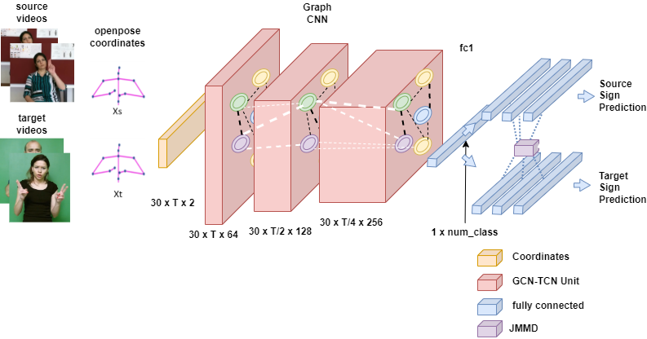}%
\caption{JMMD loss is usded to learn domain specific weights for the final layers of the network. }
\end{center}
\end{figure} 

In a different approach, rather than using straightforward entropy minimization, the \textbf{{\acf{MCC}}} approach proposed by \cite{jin2020minimum} attempts to utilize the structure of the classification output matrix to perform transfer learning. The approach minimizes classification error by minimizing pairwise class confusion among unlabeled target data within a mini-batch. The model takes as input the logit matrix and multiplies the logit matrix by its transpose to approximate a confusion matrix where correlated classes yield higher pairwise scores for each other more often. Probability rescaling and uncertainty reweighing are used to emphasize samples that are more important for classification. In contrast, category normalization is used to balance the effect of each class in a mini-batch. Finally, a loss function is calculated that maximizes the diagonal of the correlation matrix and minimizes the confusion of each class as seen in Figure 5. 



\begin{figure}[!htpb]
\label{fig:mcc}
\begin{center}
\includegraphics[width=0.9\linewidth]{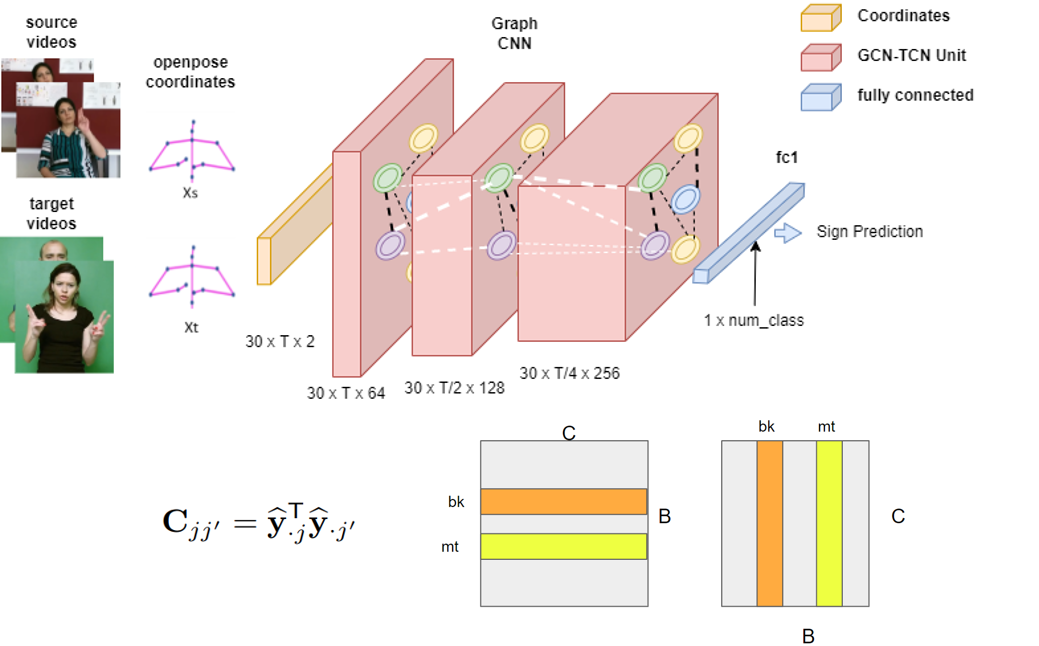}%
\caption{Calculation of minimum class confusion loss for supervised SLR. Class confusion matrix is calculated by multiplying the logit matrix by its transpose. Confident predictions increase weight of values on matrices diagonal. }
\end{center}
\end{figure}

\section{Experiments}
\label{sec:experiments}
We evaluate our methods on the \ac{BSign22k} and \ac{AUTSL} datasets. Below we introduce and detail the transfer learning subsets derived from these datasets for each experiment. 

\subsection{Datasets and Setup}
In this study, we have created a selection of similar isolated sign language signs from two publicly available sign language datasets. Going over the signs based on meaning and visual similarity, we selected 57 common signs from among 216 AUTSL and 744 BSign22k signs. For BosphorusSign22k, we also divided the dataset by number of users to create two distinct test settings with single (BSign22k\textsubscript{single}) and multi-user(BSign22k\textsubscript{shared}) target training settings. The details of each dataset are given in Table \ref{tab:dataset1}.

\begin{table*}[!htpb]

\begin{center}

\caption{Number of identical sign pairs and total number of videos from the AUTSL and BSign22k datasets.}
\resizebox{1.5\columnwidth}{!}{
\begin{tabular}{cccccc}
\toprule
 &
  \multicolumn{1}{c}{\# signs} &
  \multicolumn{1}{c}{\# train videos} &
  \multicolumn{1}{c}{\#  val.videos} &
  \multicolumn{1}{c}{\# train users} &
  \multicolumn{1}{c}{\# val. users} \\ \hline
BSign22k         & 744 & 17090 & 5452 & 4  & 2 \\ 
AUTSL            & 216 & 28139 & 3742 & 31 & 6 \\ 
BSign22k\textsubscript{shared} & 57  & 1496  & 498  & 4  & 2 \\ 
BSign22k\textsubscript{single} & 57  & 377  & 498  & 1  & 2 \\ 
AUTSL\textsubscript{shared}    & 57  & 7076  & 935  & 31 & 6 \\

\bottomrule
\end{tabular}}
\label{tab:dataset1}

\end{center}

\end{table*}

The training and validation splits of the datasets follow the same convention proposed in their identical papers in order to make experimental results compatible with other studies on those datasets \cite{ozdemir2020bosphorussign22k, sincan2020autsl}. 
In the dataset selection process, 40 of the 57 common signs are composed of signs that are performed in nearly perfect similarity. To increase domain difference between the two sets, the remaining 17 signs selected from the datasets show slight differences in hand orientation, movement direction, or the presence of additional morphemes as part of a compound sign. The dataset subsets will be made publicly available at \href{https://github.com/alpk/tid-supervised-transfer-learning-dataset/}{https://github.com/alpk/tid-supervised-transfer-learning-dataset/}. 

\begin{figure}
\begin{center}
\includegraphics[width=0.5\linewidth]{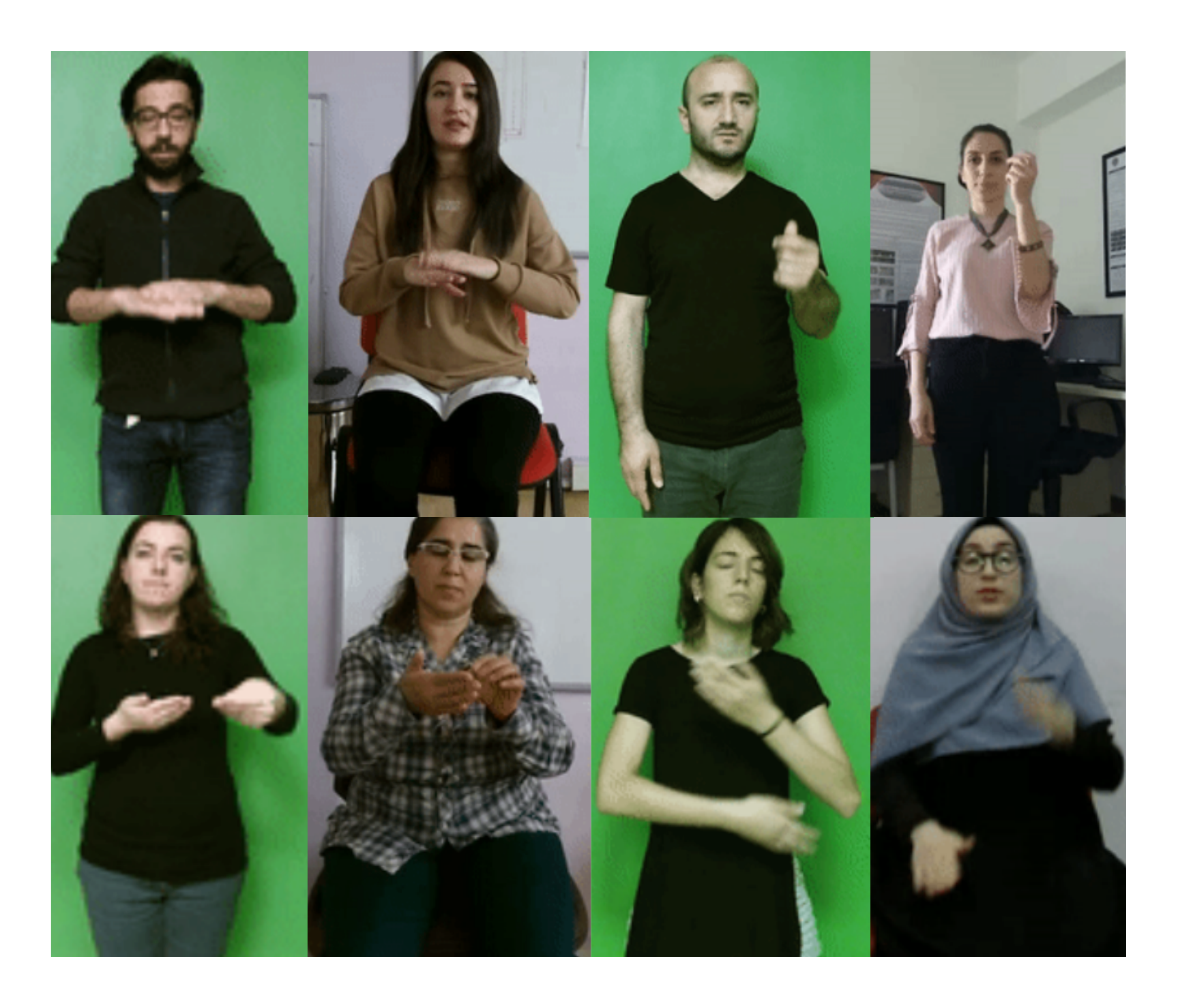}
\caption{Representative Images from the BSign22k\textsubscript{shared} (images to the left with green background) and AUTSL\textsubscript{shared} (images to the right with white \& complex backgrounds) datasets}
\end{center}
\end{figure}

\subsection{Experimental Results and Discussion}

First, we evaluate the existing datasets and baseline recognition methods to choose hyperparameters and optimize our ISLR pipeline. Next, we evaluate the cross-domain adaptation experiments on the BSign22k\textsubscript{shared} BSign22k\textsubscript{single} and AUTSL\textsubscript{shared} datasets. Then, we move on to transfer from the larger dataset, and we evaluate whether partial set transfer learning from the AUTSL dataset is beneficial.

\subsubsection{Baseline Isolated Recognition Results}
\label{sec:baseline_results}
To thoroughly evaluate the isolated recognition performance, we first evaluate the baseline recognition performance of the \ac{SL-GCN}-based recognition method without involving transfer learning. We use several different single source and finetuning-based approaches to obtain baseline results.

\begin{table}[]
\caption{Baseline accuracy values for isolated sign language recognition when no transfer learning is used during training. }
\begin{center}
    
\resizebox{\columnwidth}{!}{
\begin{tabular}{|l|c|c|c|c|}
\hline
\textbf{{}}           & \textbf{AUTSL}    & \textbf{BSign} & \textbf{AUTSL\textsubscript{shared}} & \textbf{{$BSign_{shared}$}} \\ \hline
Target Only         & 91.22 & 89.25 & 97.78   & 92.97   \\ \hline
Finetuning on AUTSL    & -                             & 89.86 & -       & 95.78   \\ \hline
Finetuning on BSign22K & 91.38                         & -     & 97.96   & -       \\ \hline
\end{tabular}}

\label{tab:methodbaseline}
\end{center}
\end{table}

In Table \ref{tab:methodbaseline}, single dataset and cross dataset accuracy results from each dataset are reported. Since we assume that no knowledge of shared classes is present, the final fully connected layer of the model, which acts as the classifier, is discarded when transferring the pretrained layer. Results show that when transferring from the larger \ac{AUTSL} dataset to the minor \ac{BSign22k} dataset, improvement is observed. Improvement is more significant on the BSign22k\textsubscript{shared} dataset, where the total number of training videos is even smaller. On the other hand, transfer from the smaller BSign22k to \ac{AUTSL} yields minimal improvement.

\subsubsection{Supervised Closed Set Transfer Learning on BSign22k}

 In this experimental setting, both the source and target datasets contain identical labels from 57 classes. Knowing the shared classes between the two datasets enables us to use supervised transfer learning approaches to facilitate better knowledge transfer between the datasets. The \textbf{target only} setting explores the accuracy of the baseline method without any input from the larger source domain. Likewise, the \textbf{source only} setting reports the accuracy of the model trained on the source domain without seeing the target domain. The two initial baseline transfer methods are the \textbf{combined} and \textbf{finetuning} approaches. In \textbf{combined}, half of the samples in each mini-batch are sampled from respective source and target domains, and the model is trained with a single cross-entropy loss. In finetuning, differing from \ref{sec:baseline_results} where we discarded the classifier fully connected layer, all layers are utilized during transfer. During finetuning, we initially freeze the feature extractor for ten epochs before unfreezing all layers and resuming training.

\begin{table}[]
\caption{Accuracy results for closed set transfer learning where only shared signs are present in training and evaluation. Each method uses AUTSL\textsubscript{shared} as source and BSign22k\textsubscript{shared} as target domains. Performance of different transfer learning methods are evaluated. }
\begin{center}
\resizebox{\columnwidth}{!}{%
\begin{tabular}{|c|c|c|cc|}
\hline
                     & \textbf{Train} & \textbf{Validation}            & \multicolumn{2}{c|}{\textbf{Train-Target}}           \\ \hline
 & \multicolumn{1}{c|}{\textbf{Source}} & \textbf{Target} & \multicolumn{1}{c|}{\textbf{BSign22k\textsubscript{single}}} & \textbf{BSign22k\textsubscript{shared}} \\ \hline
\textbf{target only} & \multicolumn{1}{c|}{}              &  BSign22k\textsubscript{shared} & \multicolumn{1}{c|}{68.28}          & 92.97          \\ \hline
\textbf{source only} & \multicolumn{1}{c|}{AUTSL\textsubscript{shared}} &          & \multicolumn{1}{c|}{65.26}          & 65.26          \\ \hline
\textbf{combined}    & \multicolumn{1}{c|}{AUTSL\textsubscript{shared}} &  BSign22k\textsubscript{shared} & \multicolumn{1}{c|}{85.11}          & 95.8           \\ \hline
\textbf{finetuning}  & \multicolumn{1}{c|}{AUTSL\textsubscript{shared}} &  BSign22k\textsubscript{shared} & \multicolumn{1}{c|}{85.34}          & 96.19          \\ \hline
\textbf{DANN}        & \multicolumn{1}{c|}{AUTSL\textsubscript{shared}} &  BSign22k\textsubscript{shared} & \multicolumn{1}{c|}{86.41}          & 96.54          \\ \hline
\textbf{MCC}         & \multicolumn{1}{c|}{AUTSL\textsubscript{shared}} &  BSign22k\textsubscript{shared} & \multicolumn{1}{c|}{86.84}          & \textbf{97.15} \\ \hline
\textbf{JAN}         & \multicolumn{1}{c|}{AUTSL\textsubscript{shared}} &  BSign22k\textsubscript{shared} & \multicolumn{1}{c|}{\textbf{88.48}} & 94.23          \\ \hline
\textbf{DSBN}        & \multicolumn{1}{c|}{AUTSL\textsubscript{shared}} &  BSign22k\textsubscript{shared} & \multicolumn{1}{c|}{84.34}          & 96.82          \\ \hline
\end{tabular}
}

\label{tab:closedsettransfer}

\end{center}
\end{table}

In Table \ref{tab:closedsettransfer}, we report on closed set transfer learning results for the transfer learning setting where AUTSL\textsubscript{shared} is set as the source task and BSign22k\textsubscript{shared} and BSign22k\textsubscript{single} subsets of the dataset are set as the target. The benefit of closed set transfer learning is higher when the number of samples in the target training set is minimal (the single-user case). In such a setting, the \ac{JAN} method surpasses the baseline finetuning accuracy with $88.48\%$. Although the baseline performance on BSign22k\textsubscript{shared} is higher, transfer learning still yields significant gains. In that setting, \ac{MCC} beats baseline finetuning and combined approaches, achieving $97.15\%$ accuracy, while methods such as \ac{JAN} and \ac{DANN} show a negligible increase.


\subsubsection{Partial Set Transfer Learning on BSign22k}

Utilizing transfer from a dataset with a larger vocabulary is also a common use case in supervised transfer learning. The source domain contains 216 signs of which 57 are common with the target dataset and 159 are additional.  Class labels for both datasets are aligned so that all methods in this setting use the same 159 class classifier. The results of these experiments are presented in  Table \ref{tab:partialsettransfer}. With all baseline and transfer learning approaches, partial-set transfer from the larger AUTSL training set surpasses the transfer results from the AUTSL\textsubscript{shared} subset. For the BSign22k\textsubscript{single} and BSign22k\textsubscript{shared} subsets, the accuracy figures reach 90.56\% and 98.63 \%, respectively with the MCC algorithm. These results show a five and one percent improvement over transfer learning baselines with combined and finetuning-based training approaches with the same experimental setting.


\begin{table}[]
\caption{Accuracy results for partial-set transfer learning are reported. Each method uses AUTSL\textsubscript{shared} as source and BSign22k\textsubscript{shared} as target domains. Performance of different Transfer learning methods are evaluated.}
\begin{center}
    
\resizebox{\columnwidth}{!}{%
\begin{tabular}{|c|c|c|cc|}
\hline
                     & \textbf{Train} & \textbf{Validation}           & \multicolumn{2}{c|}{\textbf{Train-Target}}  \\ \hline
     & \multicolumn{1}{c|}{\textbf{Source}} & \textbf{Target}  & \multicolumn{1}{c|}{\textbf{BSign22k\textsubscript{single}}} & \textbf{BSign22k\textsubscript{shared}} \\ \hline
\textbf{target only} & \multicolumn{1}{c|}{}      & BSign22k\textsubscript{shared} & \multicolumn{1}{c|}{68.28} & 92.97          \\ \hline
\textbf{source only} & \multicolumn{1}{c|}{AUTSL} &                  & \multicolumn{1}{c|}{62.79} & \textit{71.23} \\ \hline
\textbf{combined}    & \multicolumn{1}{c|}{AUTSL} & BSign22k\textsubscript{shared} & \multicolumn{1}{c|}{85.11} & 98.12          \\ \hline
\textbf{finetuning}  & \multicolumn{1}{c|}{AUTSL} & BSign22k\textsubscript{shared} & \multicolumn{1}{c|}{88.75} & 97.14          \\ \hline
\textbf{DANN}        & \multicolumn{1}{c|}{AUTSL} & BSign22k\textsubscript{shared} & \multicolumn{1}{c|}{88.75} & 98.19          \\ \hline
\textbf{MCC} & \multicolumn{1}{c|}{AUTSL}           & BSign22k\textsubscript{shared} & \multicolumn{1}{c|}{\textbf{90.56}}            & \textbf{98.63}            \\ \hline
\textbf{JAN}         & \multicolumn{1}{c|}{AUTSL} & BSign22k\textsubscript{shared} & \multicolumn{1}{c|}{90.16} & 96.72          \\ \hline
\end{tabular}
}

\label{tab:partialsettransfer}
\end{center}
\end{table}

In the baseline methods in Table \ref{tab:partialsettransfer}, the source-only method achieves lower accuracy than the same experiment with the closed set transfer learning method. As the base source, only classifier in partial-set transfer learning has more classes, it can make more mistakes. In addition, as the size of the source dataset increases, the transfer approaches \ac{DANN}, \ac{MCC} and \ac{JAN} yield scores that are closer to baseline approaches such as combined and finetuning approaches. In Table \ref{tab:fusion}, we explore the combinations of these classifiers with finetuning and each other on the BSign22k\textsubscript{shared} dataset. The fusion of these algorithms is achieved by initializing the feature extractor and classifier layers of the algorithm, freezing them for the first five epochs, and then applying respective model architecture loss combinations to a single model. Of the attempted methods, finetuning and MCC approaches yield the most significant gains. In a greedy fashion, we combined this method with several other methods, which gave us $98.8\%$ accuracy on the BSign22k\textsubscript{shared} task of the dataset.

\begin{table}[!htbp]
\caption{Top-1 Accuracy results for fusion of transfer learning methods are presented. }
\begin{center}
    
\resizebox{\columnwidth}{!}{%
\begin{tabular}{|c|cc|c|}
\hline
                        & \multicolumn{2}{c|}{\textbf{Training}}                 & \textbf{Validation} \\ \hline
                        & \multicolumn{1}{c|}{\textbf{Target}} & \textbf{Target} & BSign22k\textsubscript{shared}   \\ \hline
finetuning + DANN       & \multicolumn{1}{c|}{AUTSL}       &  BSign22k\textsubscript{shared}         & 97.14               \\ \hline
finetuning + MCC        & \multicolumn{1}{c|}{AUTSL}       &  BSign22k\textsubscript{shared}         & \textbf{98.59}      \\ \hline
finetuning + JAN        & \multicolumn{1}{c|}{AUTSL}       &  BSign22k\textsubscript{shared}         & 96.91               \\ \hline
finetuning + DSBN       & \multicolumn{1}{c|}{AUTSL}       &  BSign22k\textsubscript{shared}         & 98.19               \\ \hline
finetuning + DANN + MCC & \multicolumn{1}{c|}{AUTSL}       &  BSign22k\textsubscript{shared}         & 92.57               \\ \hline
finetuning + JAN + MCC  & \multicolumn{1}{c|}{AUTSL}       &  BSign22k\textsubscript{shared}         & 84.33               \\ \hline
finetuning + DSBN + MCC & \multicolumn{1}{c|}{AUTSL}       &  BSign22k\textsubscript{shared}         & \textbf{98.8}       \\ \hline
\end{tabular}
}

\label{tab:fusion}

\end{center}

\end{table}

\section{Conclusions}
\label{sec:conclusion}

In this paper, we establish a common sign language vocabulary subset from two publicly available Turkish Sign Language datasets and introduce experimental protocols for supervised transfer learning experiments. We believe the dataset will be a useful benchmark for testing novel supervised and unsupervised transfer learning methods for video classification.

We also propose a sign language classification method that uses graph convolutional neural networks and deep transfer learning mechanisms to improve isolated \ac{SLR} performance on under-resourced sign language datasets. We experiment with two protocols, namely, closed set transfer learning and partial-set transfer learning. Experimental results show that when shared class knowledge is present, supervised transfer learning techniques improve the performance of isolated \ac{SLR}. The observed improvement is more significant in the single-user test. In the case of closed set transfer learning, the improvements in performance over baseline methods can be attributed to the increased numbers of samples per class. Improvements with transfer methods such as MCC, DSBN, ADDA, and JAN show that improved transfer efficiency from the same class samples of a different domain also leads to a further improvement. In addition, in the case of partial-set transfer learning, the benefit observed from the addition of source data belonging to out of vocabulary signs becomes apparent as the number of signs in the training set increases and class knowledge is preserved during transfer. 

Finally, observed improvements with the fusion of MCC, finetuning, and DSBN show that applying transfer methods that focus on different aspects of neural network models such as normalization layers, initialization, and loss functions further increase the benefit that can be observed from utilizing shared class knowledge.

{\small
\bibliographystyle{ieee}
\bibliography{main}
}

\end{document}